\pdfoutput=1

\documentclass[11pt]{article}

\usepackage{ACL2023}

\usepackage{times}
\usepackage{latexsym}
\usepackage{enumitem}

\usepackage[T1]{fontenc}

\usepackage[utf8]{inputenc}

\usepackage{microtype}

\usepackage{inconsolata}

\usepackage{graphicx} 
\usepackage{amsmath,bm}
\usepackage{tabularray}
\usepackage{booktabs}
\usepackage{color}
\usepackage{diagbox}
\usepackage{multirow}
\usepackage{float}
\usepackage{stfloats}

\title{Improving Retrieval Augmented Open-Domain Question-Answering with Vectorized Contexts}

\author{Zhuo Chen$^{1}$, Xinyu Wang$^{2}$\thanks{\hspace{0.3em} Corresponding author}, Yong Jiang$^{2*}$, Pengjun Xie$^{2}$, Fei Huang$^{2}$, Kewei Tu$^{1*}$ \\
        $^{1}$School of Information Science and Technology, ShanghaiTech University \\
        $^{1}$Shanghai Engineering Research Center of Intelligent Vision and Imaging \\
        $^{2}$Institute for Intelligent Computing, Alibaba Group \\
        \texttt{\{chenzhuo,tukw\}@shanghaitech.edu.cn} \\
        \texttt{\{tomas.wxy,yongjiang.jy,chengchen.xpj\}@alibaba-inc.com}
        }

\begin{document}

\maketitle

\begin{abstract}

In the era of large language models, applying techniques such as Retrieval Augmented Generation can better address Open-Domain Question-Answering problems. Due to constraints including model sizes and computing resources, the length of context is often limited, and it becomes challenging to empower the model to cover overlong contexts while answering questions from open domains. This paper proposes a general and convenient method to cover longer contexts in Open-Domain Question-Answering tasks. 
It leverages a small encoder and cross-attention mechanism and effectively encodes contexts. With our method, the original language models can cover several times longer contexts while keeping the computing requirements close to the baseline. Our experiments demonstrate that after fine-tuning, there is improved performance across two held-in datasets, four held-out datasets, and also in two In Context Learning settings. Our code will be released at {\url{https://github.com/Alibaba-NLP/Vec-RA-ODQA}}.
\end{abstract}

\section{Introduction}

Transformer-based \cite{vaswani2017attention} architectures with pre-training on large corpus have become popular in recent Natural Language Processing research \cite{brown2020language, workshop2022bloom, chowdhery2023palm}. An increasing number of Natural Language Processing (NLP) tasks need to process long contexts such as Open-Domain Question Answering (ODQA) with Retrieval Augmented Generation (RAG) \cite{lewis2020retrieval, izacard2020leveraging, gu2018search}. However, the fine-tuning and inference stages in downstream tasks are still constrained by the input length, e.g., 2048 tokens for Bloomz \cite{muennighoff2022crosslingual} and Llama-1 \cite{touvron2023llama}.
\begin{figure}[tb]
\centering
\includegraphics[width=1.0\linewidth]{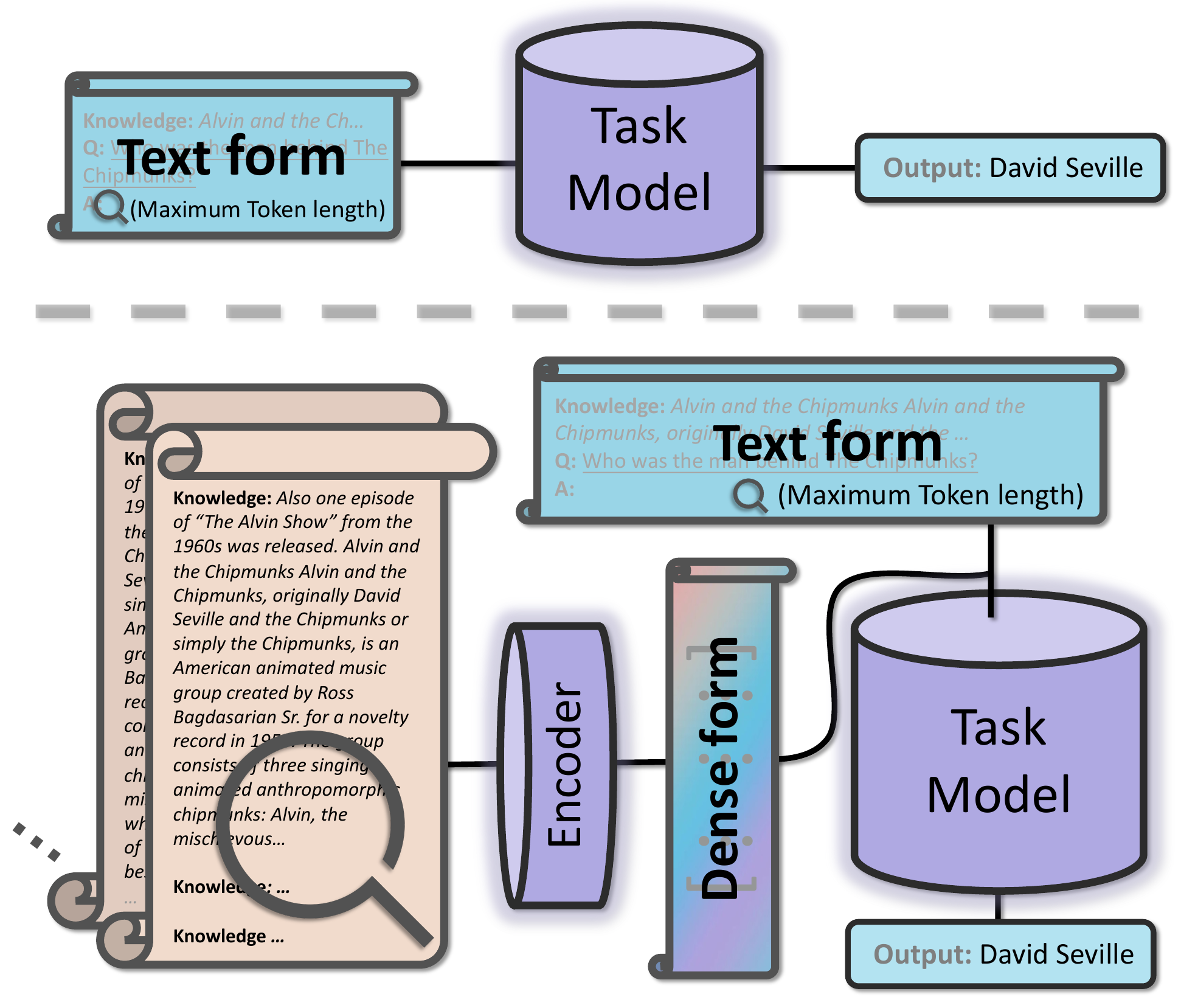}
\caption{A comparison of our method (lower) and retrieval augmented ODQA without vectorization (upper). In the upper part, limited retrieved contexts are processed by the task model to finish the task. The lower part illustrates our method in which an encoder is incorporated to encode overlong retrieved contexts.}
\label{frame}
\end{figure}

With RAG, the input can easily surpass the maximum length the model can handle and it becomes challenging for the model to perform both fine-tuning and inference on overlong contexts. Moreover, in the in-context learning (ICL) \cite{dong2022survey, kim2022self} setting, the context will be much longer together with retrieved contexts. In such cases, the demand for the model to handle longer input text significantly increases. 

To enable the model to cover longer context during both fine-tuning and inference stages, this paper proposes a method that leverages a 100 million-level encoder model in downstream ODQA tasks with a 1 billion-level language model as illustrated in the lower part of Fig.~\ref{frame}. With our method, the length of context that the model can cover increases from 2k (in text form) to a maximum of 10k (in dense form, which is condensed by the encoder). Experiments are designed under three settings to validate the effectiveness of our method. In the experiments, we first fine-tune the model, optionally including the encoder, on two popular ODQA datasets with retrieved contexts and evaluate our method in held-in, held-out, and ICL settings. Experimental results show that our method outperforms the baseline, which is fine-tuned on data of length 2k, in all three settings.

\begin{figure}
\centering
\includegraphics[width=1.0\linewidth]{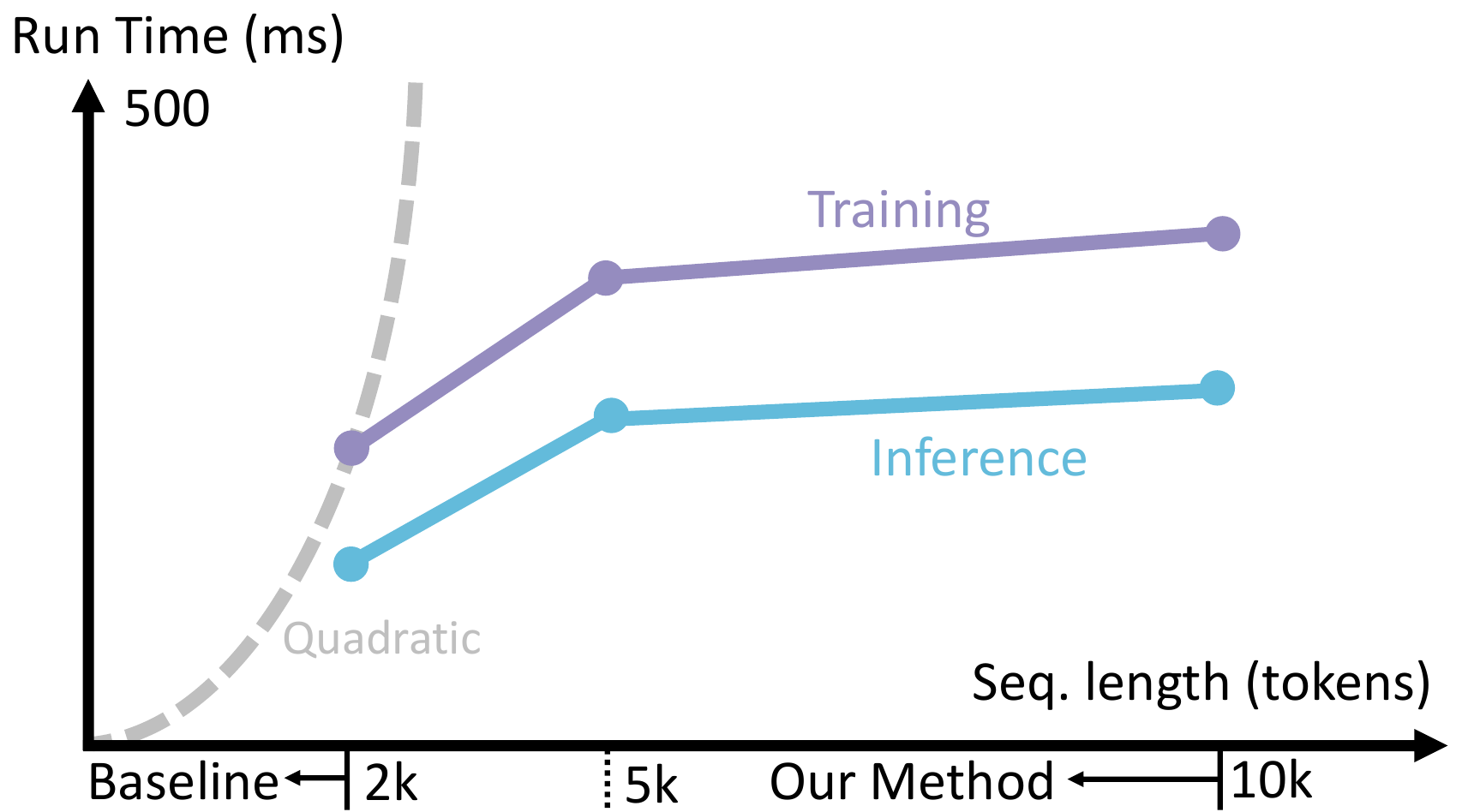}
\caption{Speed illustration. Run time is measured on a single A100 GPU and the \texttt{batch size} is set to 1 for all curves. "2k" on the horizontal axis represents the baseline model's run time to train or infer on data of length 2k. "5k" and "10k" correspond to two variants of our method that can cover at most 5k and 10k tokens when training and inferring. Training time measures the average over five consecutive training steps. Inference time measures the average over five consecutive generation steps. Specifically, we measure the execution duration of functions \texttt{Trainer.training\_step} and \texttt{model.generate} based on \href{https://www.huggingface.co/}{\texttt{huggingface}}. }
\label{eff}
\end{figure}

Regarding the speed of our method, we measure the run time of each training and inference step. Compared with work that compresses the contexts with the original task model \cite{chevalier2023adapting}, which requires techniques to reduce the computation graph during backpropagation, we employ a 10x smaller model to perform the encoding of excessive texts, so a complete gradient descent procedure can be kept. 
To sum up, our contributions are as follows:
\begin{enumerate}[leftmargin=*,noitemsep,topsep=1pt]
    \item We propose a method that incorporates a small encoder model for excessively long context encoding by applying cross-attention mechanism with the original task model.
    \item We evaluate our method in two held-in, four held-out, and two ICL settings after being fine-tuned on two ODQA datasets and obtain improved performance.
    \item The computing resource requirements of our method are consistent with those of the baseline and the run time remains competitive.
\end{enumerate}

\section{Method}
\label{method}

\begin{figure}[tb]
\centering
\includegraphics[width=0.78\linewidth]{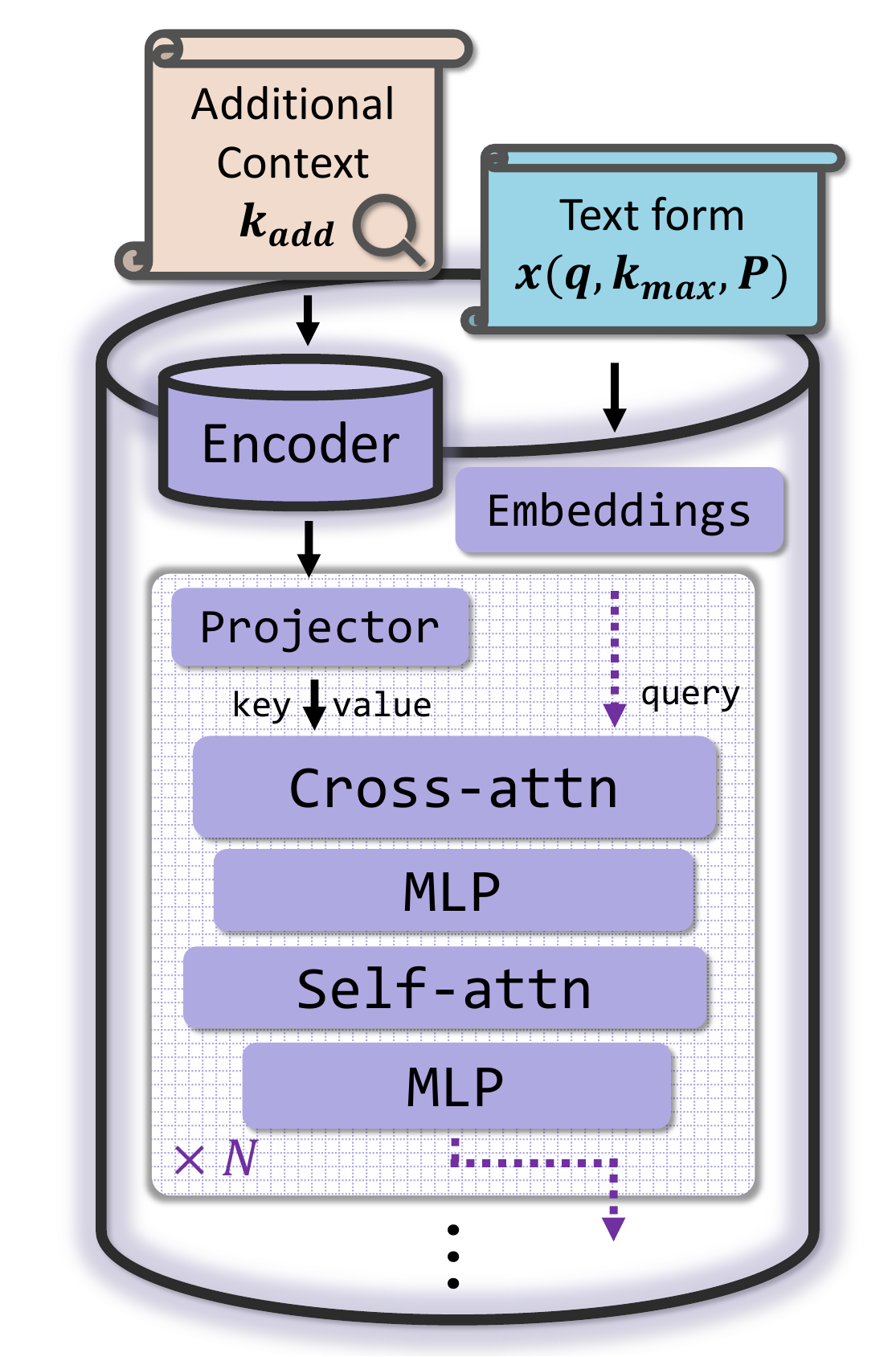}
\caption{Method illustration of model architecture (purple blocks) and data flows (along black/purple arrows). The purple dashed arrows mean that the output of \texttt{MLP} module will be the "query" to the next layer of \texttt{Cross-attn} module. $\times N$ means that the modules with dotted backgrounds are repeated with multiple layers in the task model.}
\label{method_fig}
\end{figure}

\subsection{Background} 

Consider an example query $\bm{q}$ with gold answer $\bm{a}$ and independent $C$ pieces of corresponding context information $\bm{k} = \{\bm{k_1}, \bm{k_2}, ..., \bm{k_C}\}$, with each being a sequence of tokens, where $\bm{k}$ is retrieved by some retriever from a given corpus\footnote{Refer to Sec.~\ref{exp_set} for detailed definition of corpus and retriever in our experiments.}
$$\bm{k} = \text{Retriever}(\bm{q}, \text{corpus})$$
Ideally, the $C$ retrieved contexts contain the knowledge needed to answer $\bm{q}$ correctly, but there may also be noise. Given a decoder model $Dec$ parameterized by $\theta$, the output sequence $\bm{y}$ is usually modeled by
$$P_\theta(\bm{y} | \bm{q}, \bm{k_{max}}, \bm{P}) = Dec(\bm{y} | \bm{q}, \bm{k_{max}}, \bm{P})$$
where $\bm{k_{max}} = \{\bm{k_1}, \bm{k_2}, ..., \bm{k_m}\} \in \bm{k}, m < C$. $m$ refers to the number of contexts that reach the model's throughput. $\bm{P}$ stands for the prompts that connect related content\footnote{The forms of $\bm{P}$ vary with different settings, and there will be detailed definitions in Sec.~\ref{exp_set}.}. Given the model, $\bm{k_{max}}$ is usually a subset of $\bm{k}$ because the maximum length of contexts is often constrained by the model's throughput or computing resources, and  

During training, we aim to maximize the term $P_\theta(\bm{a} | \bm{q}, \bm{k_{max}}, \bm{P})$, and formalize the ODQA problem as a language modeling task. Specifically, for a query $\bm{q}$, its gold answer $\bm{a}$ and contexts $\bm{k_{max}}$, they are connected linguistically with proper prompts $\bm{P}$, together denoted as an input sequence $\bm{x}(\bm{q}, \bm{a}, \bm{k_{max}}, \bm{P}) = \{x_1,x_2, ...\}$. Then we aim to minimize the language modeling loss over the set $\mathcal{D}$ of all training examples:

\begin{align}\label{objective}
\begin{split}
L_\theta(\mathcal{D}) = - \sum\limits_{ \bm{x}(\bm{q}, \bm{a}, \bm{k_{max}}, \bm{P}) \in \mathcal{D}} &\sum\limits_{i} \\[4pt] 
log(P_\theta(x_i&|x_{<i}))
\end{split}
\end{align}

\subsection{Encoding and Cross-Attention}
\label{enc_and_cros}
We propose a method that can utilize additional contexts $\bm{k_{add}} = \{\bm{k_{m+1}}, \bm{k_{m+2}}, ...\}$ several times longer than $\bm{k_{max}}$. First, we introduce an encoder parameterized by $\phi$. Then we apply cross-attention with the original task model and introduce a projector, a cross-attention module and a Multi-Layer Perceptron (MLP) in each layer, together denoted the parameters as $\pi$. Denote $\omega=\{\phi, \pi, \theta\}$ as all the parameters in our model. On the whole, our method models the output $\bm{y}$ by an encoder-decoder model $Enc$-$Dec$
\vspace{1pt}
\begin{align*}
&Q_{\omega}(\bm{y} | \bm{q}, \bm{k_{max}}, \bm{P}, \bm{k_{add}}) \quad \\[4pt]
= ~ &Enc\text{-}Dec(\bm{y} | \bm{q}, \bm{k_{max}}, \bm{P}, \bm{k_{add}}) \quad
\end{align*}
During training, inputs $\bm{x} (\bm{q}, \bm{a}, \bm{k_{max}}, \bm{P})$ are embedded by the original task model's embedding layer $Emb$
\vspace{1pt}
\begin{align*}
\bm{h_{q}} = Emb(\bm{x} (\bm{q}, \bm{a}, \bm{k_{max}}, \bm{P}))
\end{align*}
and each of the additional contexts $\bm{k_i}$ in  $\bm{k_{add}}$ is encoded by the encoder $Enc$
\vspace{1pt}
\begin{align*}
\bm{h_{add}^{(i)}} = Enc(\bm{k_i})
\end{align*}
Note that the length of encoding from the encoder is flexible practically and we compress each $\bm{k_i}$ into one vector. Following the output of the encoder, a projector $Proj$ is used to align the high-dimensional hidden spaces between the encoder and task model in each layer
\vspace{1pt}
\begin{align*}
\bm{h_{kv}} = Proj(\bm{h_{add}})
\end{align*}
where $\bm{h_{add}}$ is concatenated of all $\bm{h_{add}^{(i)}}$ calculated from last step. Each layer of the task model is assigned to an independent projector as different layers may learn different representations.

In each layer, to incorporate the information stored in $\bm{k_{add}}$ we add a cross-attention module, where representations of additional contexts $\bm{h_{kv}}$ serve as "key" and "value", followed by an MLP. In the first layer, the embeddings of original input $\bm{h_{q}}$ act as "query", and in the rest of the layers output $\bm{h_{q}^{'}}$ from the previous layer act as "query" ($\bm{h_{q}^{'}}$ will be defined later).
\vspace{1pt}
\begin{align*}
\bm{h_{c}} &= Cross\text{-}attn(\bm{h_{q}}/\bm{h_{q}^{'}}, \bm{h_{kv}}) \\[1pt]
\bm{h_{m}} &= MLP(\bm{h_c}) 
\end{align*}
$Cross\text{-}attn(\bm{h_{q}}, \bm{h_{kv}})$ is calculated as follows
\begin{align*}
Q &= W^Q \bm{h_{q}} \\[4pt]
K, V &= W^K \bm{h_{kv}}, W^V \bm{h_{kv}} \\[4pt]
o &= softmax(\frac{QK^T}{\sqrt{d_k}})V \\[4pt]
\bm{h_{c}} &= W^O o
\end{align*}
where $W^Q, W^K, W^V, W^O$ refer to weight matrices and $d_k$ refers to the dimension of each attention head.
Then the output of cross-attention and MLP is normally processed by a self-attention and another MLP module. The output acts as "query" input to the cross-attention module in the next layer. 
\begin{align*}
\bm{h_{q}^{'}} &= MLP( Self\text{-}attn(\bm{h_{m}}) )
\end{align*}

At last, the output of the last layer is expanded to the vocabulary-size dimension to predict the next token (not shown in Fig.~\ref{method_fig} for simplicity), and we aim to maximize the probability
$$Q_{\omega}(\bm{a} | \bm{q}, \bm{k_{max}}, \bm{P}, \bm{k_{add}})$$

Consistent with the setup mentioned before, to maximize term  $Q_{\omega}(\bm{a} | \bm{q}, \bm{k_{max}}, \bm{P}, \bm{k_{add}})$, we turn it into minimizing the language modeling loss
\begin{equation}\label{our_objective}
\begin{split}
J_{\omega}(\mathcal{D}) = - \sum\limits_{ \bm{x}(\bm{q}, \bm{a}, \bm{k_{max}}, \bm{P}), \bm{k_{add}} \in \mathcal{D}} &\sum\limits_{i} \\[4pt]
log (Q_{\omega}(x_i&|x_{<i}, \bm{k_{add}}))
\end{split}
\end{equation}

\subsection{ICL Setting}
\label{icl_setting_sec}

Our method can also be applied to ICL settings. Based on the aforementioned setup, we denoted ICL samples as $\bm{l_{max}} = \{\bm{l_1}, \bm{l_2}, ..., \bm{l_m}\}$, with each $\bm{l_i}$ composed of another pair of query and answer. We optimize objective \ref{our_objective_icl} below on data where each $\bm{l_i}(\bm{q^{'}}, \bm{a^{'}})$ refers to only query-answer ICL samples (without context) and $\bm{q^{'}}$ $\bm{a^{'}}$ refer to another query-answer pair:
\begin{equation}\label{our_objective_icl}
\begin{split}
J^{'}_{\omega}(\mathcal{D}) = - \sum\limits_{ \bm{s}(\bm{q}, \bm{a}, \bm{l_{max}}, \bm{P}), \bm{k_{add}} \in \mathcal{D}} &\sum\limits_{i} \\[4pt]
log(Q^{'}_{\omega}(s_i&|s_{<i}, \bm{k_{add}}))
\end{split}
\end{equation}
$\bm{s}=\{s_1, s_2, ... \}$ refers to the inputs composed of $(\bm{q}, \bm{a}, \bm{l_{max}}, \bm{P})$ and $Q^{'}$ shares a similar definition to $Q$ in objective \ref{our_objective}. Additional contexts $\bm{k_{add}} $ are utilized in the same way as in Sec.~\ref{enc_and_cros} by performing encoding, cross-attention, etc.

\subsection{Training}
\label{training_sec}

Theoretically, training processes stated in Sec.~\ref{enc_and_cros} all remain differentiable and thus all the parameters can be optimized via normal gradient descent w.r.t. objective \ref{our_objective}. Note that the parameters $\phi$ of the encoder can be initialized from a well-pre-trained model on a large scale corpus and the pre-trained parameters possess good performance in many downstream tasks based on text encoding. However, the parameters in the projector module are randomly initialized. Thus at the start of the training, according to the chain rule, the gradients to the whole encoder will be random as well, which poses a risk of breaking the encoding utility of the encoder. This intuition proves to be true in our experiments. 

Therefore, we design two strategies of training:

\begin{enumerate}
    \item Directly freeze parameters $\phi$ and make parameters $(\pi, \theta)$ trainable during the whole training process.
    \item In the first few training steps (e.g., one epoch), $\phi$ is kept frozen to prevent random gradients from breaking its well-pre-trained parameters. After that, $\phi$ is optimized w.r.t. objective \ref{our_objective} together with the other modules $(\pi, \theta)$.
\end{enumerate}

\section{Experiment}

\subsection{Experiment settings}
\label{exp_set}

\paragraph{Data}

\begin{table}[tb]
\renewcommand\arraystretch{1.05}

\centering
\scalebox{0.8}{
\begin{tabular}{cl} 
\toprule
\textbf{Settings} & \multicolumn{1}{c}{\textbf{Data format}} \\ 
\midrule
\midrule
\begin{tabular}[c]{@{}c@{}}\textbf{Held-in }\\\textbf{Held-out}\end{tabular} & \begin{tabular}[c]{@{}l@{}}\textcolor[rgb]{0.753,0.753,0.753}{\textbf{Answer the question:}}\\\textcolor[rgb]{0.753,0.753,0.753}{\textbf{Knowledge}:} \{context $\bm{k_1}$~\}\\...\{context~$\bm{k_m}$\}.\\\textcolor[rgb]{0.753,0.753,0.753}{\textbf{Q}:} Who got the first nobel prize in physics\\\textcolor[rgb]{0.753,0.753,0.753}{\textbf{A}:}\end{tabular} \\ 
\midrule
\begin{tabular}[c]{@{}c@{}}\textbf{ICL format}\\\textbf{w/ contexts}\end{tabular} & \begin{tabular}[c]{@{}l@{}}\textcolor[rgb]{0.753,0.753,0.753}{\textbf{Answer the following questions based }}\\\textcolor[rgb]{0.753,0.753,0.753}{\textbf{on the Knowledge:}}\\\textcolor[rgb]{0.753,0.753,0.753}{\textbf{Knowledge}:} \{context $\bm{k_1}^{'}$\}\\\textcolor[rgb]{0.753,0.753,0.753}{\textbf{Q}:} Who developed the first printing press \\in 1430s\\\textbf{\textcolor[rgb]{0.753,0.753,0.753}{ A}}\textcolor[rgb]{0.753,0.753,0.753}{:} Johannes Gutenberg\\ ...(\textbf{\textcolor[rgb]{0.753,0.753,0.753}{Knowledge}}\textcolor[rgb]{0.753,0.753,0.753}{: }... \textcolor[rgb]{0.753,0.753,0.753}{\textbf{Q}:} ... \textcolor[rgb]{0.753,0.753,0.753}{\textbf{A}:} ...)\\\textcolor[rgb]{0.753,0.753,0.753}{\textbf{Knowledge}:~}\{context $\bm{k_1}^{''}$\}\\\textcolor[rgb]{0.753,0.753,0.753}{\textbf{Q}:} Who got the first nobel prize in physics\\\textcolor[rgb]{0.753,0.753,0.753}{\textbf{A}:}\end{tabular} \\ 
\midrule
\begin{tabular}[c]{@{}c@{}}\textbf{ICL format}\\\textbf{w/o contexts}\\\textbf{(Sec.~\ref{icl_setting_sec})}\end{tabular} & \begin{tabular}[c]{@{}l@{}}\textcolor[rgb]{0.753,0.753,0.753}{\textbf{\textbf{Answer the following questions}}}\textcolor[rgb]{0.753,0.753,0.753}{:}~\\\textcolor[rgb]{0.753,0.753,0.753}{\textbf{\textbf{Q}}:}~Who developed the first printing press\\in 1430s\\\textbf{\textbf{\textcolor[rgb]{0.753,0.753,0.753}{A}}}\textcolor[rgb]{0.753,0.753,0.753}{:}~Johannes Gutenberg\\(\textcolor[rgb]{0.753,0.753,0.753}{\textbf{\textbf{Q}}:}~...~\textcolor[rgb]{0.753,0.753,0.753}{\textbf{\textbf{A}}:}~...)...\\\textcolor[rgb]{0.753,0.753,0.753}{\textbf{\textbf{Q}}:}~Who got the first nobel prize in physics\\\textcolor[rgb]{0.753,0.753,0.753}{\textbf{\textbf{A}}:}\end{tabular} \\ 
\midrule
\begin{tabular}[c]{@{}c@{}}\textbf{Additional}\\\textbf{Contexts}\end{tabular} & \begin{tabular}[c]{@{}l@{}}\{context $\bm{k_{m+1}}$\};~\\\{context $\bm{k_{m+2}}$\};\\...\end{tabular} \\
\bottomrule
\end{tabular}}
\caption{Examples of data format. \textbf{\textcolor[rgb]{0.753,0.753,0.753}{Gray tokens}} refer to prompts $\bm{P}$ mentioned in Sec.~\ref{method} and the context is omitted here.}
\label{data_format}
\end{table}

To evaluate our method, we first fine-tune our model on two ODQA datasets separately, TriviaQA \cite{joshi-etal-2017-triviaqa} and Natural Questions (NQ) \cite{kwiatkowski-etal-2019-natural}. Besides evaluating our method on the held-in data, we also evaluate four held-out data, namely CommonsenseQA \cite{talmor2019commonsenseqa}, SQuAD2.0 \cite{rajpurkar2016squad}, Webquestions \cite{berant-etal-2013-semantic} and ComplexWebQuestions \cite{talmor2018web}. Specifically, samples in CommonsenseQA dataset are formulated as multi-choice problems, and we evaluate the performance in both multi-choice and sequence-to-sequence formats. Refer to App.~\ref{comqa_app} for the detailed format. 

Format of input $\bm{x}$ in Sec.~\ref{enc_and_cros} is formulated as "Held-in Held-out" format in Table~\ref{data_format}, and we evaluate the model's performance on samples of ICL format with context. Format of input $\bm{s}$ in Sec.~\ref{icl_setting_sec} is formulated as "ICL format w/o contexts" in Table~\ref{data_format}.

Additional contexts $\bm{k_{m+1}}, \bm{k_{m+2}}$ are encoded by the encoder separately and independently without prompts. The forms of prompts $\bm{P}$ defined previously are shown in gray tokens in Table~\ref{data_format}.

\paragraph{Retriever}
For contexts of the datasets TriviaQA and NQ, we utilize those collected by \citet{karpukhin2020dense}, which are collected with BM25 \cite{robertson2009probabilistic} and Dense Passage Retrieval techniques. For contexts of the four held-out datasets, we follow \citet{izacard2022few} and \citet{shi2023replug} and use Contriver \cite{izacard2021unsupervised} as our retriever. Contexts $\bm{k}$ are retrieved from Wikipedia dump dated December 20, 2018, the version released by \citet{karpukhin2020dense}.

\paragraph{Baseline}
Recent decoder-only models like Bloomz \cite{muennighoff2022crosslingual} and GPTs \cite{radford2019language, achiam2023gpt} have shown good performance in generation-like tasks, and we use \textit{Bloomz-1b7}\footnote{\url{https://huggingface.co/bigscience/bloomz-1b7}} for the task model $\theta$. When fine-tuning the baseline model, inputs are constructed according to the "Held-in Held-out" setting as stated in Table~\ref{data_format}. The length of the input is extended to utilize as many contexts as possible, consistent with the maximum input length (2k) of the model while doing pre-training \cite{workshop2022bloom}.

Additionally, note that the context information $\bm{k_{max}}$ provided in the inputs is ranked from best to worst based on Dense Retrieval \cite{karpukhin2020dense}, which means the baseline we adopt is rather stronger than randomly providing as many contexts as possible without considering the quality. The baseline can be seen as a model fine-tuned on the most relevant contexts incorporating reranking techniques \cite{karpukhin2020dense, khalifa-etal-2023-shot}.

\paragraph{Initialization and Training Settings}

Weights of popular pre-trained encoder models like BERT \cite{devlin2018bert} should be good initialization for the encoder $\phi$ and thus we adopt \textit{BERT-base-uncased}\footnote{\url{https://huggingface.co/bert-base-uncased}} for initialization of $\phi$. Parameters of attention and MLP modules are also adapted from \textit{Bloomz-1b7}. To keep the encoding process efficient, we use a simple \texttt{Linear} module as the projector that is randomly initialized and fine-tuned to align the hidden dimension of 768 (\textit{BERT-base-uncased}) to 2048 (\textit{Bloomz-1b7}).

In our experiment, we use BERT to independently encode additional contexts on 10 or 20 contexts, which can cover approximately 5k to 10k additional context tokens. Then the hidden states of the \textit{[CLS]} token are concatenated and fed-forward to subsequent modules as illustrated in Fig.~\ref{method_fig}. For both the baseline and our method, we evaluate the model checkpoint with the lowest language modeling loss on the development set and report the Exact Match (EM) metric.

As discussed in Sec.~\ref{training_sec}, there are mainly two choices of training strategies of which parts of our proposed model are optimized. We experiment with both strategies and report the results of the "frozen encoder" setting in Sec.~\ref{main_result} and the "training encoder" setting in Sec.~\ref{enc_train} respectively. 

\paragraph{Hyperparameters}
We list important hyperparameters in our experiments in Table~\ref{hyper}.

\renewcommand\arraystretch{1.0}

\begin{table}[t]
\centering
\small
\refstepcounter{table}
\begin{tabular}{ll} 
\toprule
Learning Rate & 2e-5 \\ 
\midrule
Optimizer & AdamW \\ 
\midrule
Lr scheduler & cosine \\
\midrule
Warmup ratio & 0.03 \\
\midrule
FP 16 & True \\
\midrule
FP 16 eval & True \\
\midrule
Globa batch size & 8 \\ 
\midrule
Save steps & 4000 \\
\midrule
Eval steps & 4000 \\
\midrule
Max epochs & 4 \\
\midrule
GPU name & NVIDIA A100-SXM 80G \\
\bottomrule
\end{tabular}
\caption{Hyperparameters}
\label{hyper}
\end{table}

\subsection{Main Results}
\label{main_result}

\renewcommand\arraystretch{1.3}

\begin{table*}[tb]
\tabcolsep=0.08cm
\centering
\scalebox{0.79}{
\begin{tabular}{l|l|ccccccccccccc} 
\toprule
\multicolumn{2}{c|}{\multirow{2}{*}{\textbf{Train \textbackslash{} Evaluate}}} & \multicolumn{2}{c}{\textbf{TriviaQA}} & \multicolumn{2}{c}{\textbf{NQ}} & \multicolumn{2}{c}{\textbf{Com.QA }\textit{test}} & \textbf{SQuAD} & \textbf{Web.Q} & \textbf{Comp.Q} & \multicolumn{2}{c}{\textbf{Triviaqa (ICL)}} & \multicolumn{2}{c}{\textbf{NQ (ICL)}} \\
\multicolumn{2}{c|}{} & \textit{dev} & \textit{test} & \textit{dev} & \textit{test} & \textit{choice} & \textit{seq2seq} & \textit{test} & \textit{test} & \textit{test} & \textit{dev} & \textit{test} & \textit{dev} & \textit{test} \\ 
\midrule
\midrule
\multirow{3}{*}{\textbf{TriviaQA}} & baseline & 45.740 & 46.203 & 14.868 & 16.288 & 17.199 & 2.785 & 10.191 & 9.524 & 4.490 & 31.764 & 31.857 & 8.999 & 9.058 \\
 & + 5k & 47.686 & 47.742 & 17.506 & 19.307 & \textbf{19.328} & \textbf{3.194} & \textbf{12.684} & 10.053 & 5.513 & 32.341 & 32.034 & 10.677 & 11.136 \\
 & + 10k & \textbf{47.901} & \textbf{48.245} & \textbf{18.465} & \textbf{19.529} & 17.363 & 2.539 & 12.667 & \textbf{11.111} & \textbf{6.024} & \textbf{34.027} & \textbf{34.235} & \textbf{11.671} & \textbf{11.801} \\ 
\midrule
\multirow{3}{*}{\textbf{NQ}} & baseline & 42.809 & 43.976 & 37.159 & 37.978 & 19.410 & 4.095 & 21.199 & 14.815 & \textbf{13.498} & 35.521 & 35.967 & 19.242 & 21.136 \\
 & + 5k & 43.669 & 44.657 & 37.01 & 38.698 & 19.656 & 4.095 & 22.724 & 15.344 & 13.214 & 35.883 & 35.985 & 19.265 & 21.413 \\
 & + 10k & \textbf{44.189} & \textbf{45.107} & \textbf{37.581} & \textbf{39.114} & \textbf{21.294} & \textbf{4.423} & \textbf{22.918} & \textbf{15.873} & 13.413 & \textbf{36.381} & \textbf{36.569} & \textbf{19.447} & \textbf{21.662} \\
\bottomrule
\end{tabular}}
\caption{Main results of performance with frozen encoder on held-in, held-out and ICL settings. \textbf{Boldface} marks the best results in each setting. Com.QA refers to CommonsenseQA. Web.Q refers to WebQuestions. Comp.Q refers ComplexWebQuestions. TriviaQA (ICL) and NQ (ICL) show the results evaluated on ICL setting where the data is formed as illustrated in Table~\ref{data_format} ICL.}
\label{main_result_table}
\end{table*}

We present our main result of the first training strategy discussed in Sec.~\ref{training_sec} in Table~\ref{main_result_table}. Upon fine-tuning on two datasets and evaluating on three (held-in, held-out and ICL) settings, our method achieves performance superior to that of the baseline in five out of six settings, except for one setting on one dataset. 

In held-in settings (training on TriviaQA/NQ and evaluating on TriviaQA/NQ), our model consistently demonstrates superior performance relative to the baseline. Moreover, it demonstrates stable improved performance as more contexts are encoded by our method, showing the potential of our model to encode even longer contexts. 

In held-out settings, our method outperforms the baseline in all the datasets after being fine-tuned on TriviaQA and outperforms three of four datasets after being fine-tuned on NQ, suggesting the general applicability of our method. From the "Com.QA \textit{choice}" setting we can see that though our model is not trained to answer multi-choice questions, it performs better in selecting choices than baseline.

In the last two columns TriviaQA (ICL) and NQ (ICL), we evaluate whether the optimized model can generalize to a similar ICL setting. Specifically, with optimized parameter $\omega^*$ after fine-tuning objective \ref{our_objective} we evaluate how well we can model 
$Q_{\omega^*}(\bm{a} | \bm{q}, \bm{l_{max}}, \bm{P}, \bm{k_{add}})$
where each $\bm{l_i}(\bm{q^{'}}, \bm{a^{'}}, \bm{k^{'}})$ is an ICL sample composed of another query, context and answer. Surprisingly, we obtain a similar improved performance to the held-in setting. Steadily improved performance indicates that the training method we adopt is robust, maintaining both the encoder and decoder's efficacy in retrieving useful information while the evaluation data format diverges from the training data.

In summary, from the results presented in Table~\ref{main_result_table}, it is observable that in comparison with the baseline, employing our method to encode a greater volume of retrieval information offers a predominantly positive enhancement to the model's performance across various settings, including held-in, held-out, and ICL.
\section{Analysis}
\label{analysis}
In this section, we present the results of three analytical experiments. The first one shows the result of the other training strategy discussed in Sec.~\ref{training_sec}. The second shows the evaluation results of optimizing objective \ref{our_objective_icl}. The third shows the effectiveness of our method in a more challenging setting.

\subsection{Encoder Training}
\label{enc_train}

In our experiments, we first try optimizing the encoder $\phi$ with the other parameters $(\pi, \theta)$ from the very beginning of the training process. Results turn out to verify our anticipation: newly introduced random parameters (the projector) easily mess up with the parameters in the encoder, consequently undermining its capability to encode information and resulting in worse performance than baseline. 

Here we evaluate the training strategy we proposed in Sec.~\ref{training_sec} that aims to fix this problem. The encoder is optimized after several training steps, and in our experiment, we set it to one epoch. Besides, the parameters in the cross-attention module are initialized by those in the pre-trained self-attention module to minimize the amount of randomly initialized parameters. 

\begin{table*}[tb]
\tabcolsep=0.08cm
\centering
\scalebox{0.79}{
\begin{tabular}{l|l|ccccccccccccc} 
\toprule
\multicolumn{2}{c|}{\multirow{2}{*}{\textbf{Train \textbackslash{} Evaluate}}} & \multicolumn{2}{c}{\textbf{TriviaQA}} & \multicolumn{2}{c}{\textbf{NQ}} & \multicolumn{2}{c}{\textbf{Com.QA }\textit{test}} & \textbf{SQuAD} & \textbf{Web.Q} & \textbf{Comp.Q} & \multicolumn{2}{c}{\textbf{Triviaqa (ICL)}} & \multicolumn{2}{c}{\textbf{NQ (ICL)}} \\
\multicolumn{2}{c|}{} & \multicolumn{1}{c}{\textit{dev}} & \textit{test} & \textit{dev} & \textit{test} & \textit{choice} & \textit{seq2seq} & \textit{test} & \textit{test} & \textit{test} & \textit{dev} & \textit{test} & \textit{dev} & \textit{test} \\ 
\midrule
\midrule
\multirow{3}{*}{\textbf{TriviaQA}} & baseline & 45.740 & 46.203 & 14.868 & 16.288 & \textbf{17.199} & 2.785 & 10.191 & 9.524 & 4.490 & \textbf{31.764} & \textbf{31.857} & \textbf{8.999} & \textbf{9.058} \\
 & + 5k & \textbf{48.082} & \textbf{47.803} & 15.896 & 16.898 & 14.333 & \textbf{3.112} & 10.755 & \textbf{11.640} & 4.945 & 16.646 & 16.645 & 6.178 & 6.676 \\
 & + 10k & 47.980 & 47.750 & \textbf{16.079} & \textbf{17.008} & 15.807 & 2.867 & \textbf{10.823} & 11.111 & \textbf{5.058} & 16.103 & 16.300 & 6.383 & 7.008 \\ 
\midrule
\multirow{3}{*}{\textbf{NQ}} & baseline & 42.809 & 43.976 & 37.159 & 37.978 & \textbf{19.410} & \textbf{4.095} & 21.199 & 14.815 & \textbf{13.498} & 35.521 & \textbf{35.967} & \textbf{19.242} & \textbf{21.136} \\
 & + 5k & \textbf{43.397} & \textbf{44.524} & \textbf{37.387} & 39.03 & 15.889 & \textbf{4.095} & 22.092 & 14.286 & 12.958 & 33.993 & 34.341 & 17.951 & 19.640 \\
 & + 10k & 43.284 & 44.047 & 37.205 & \textbf{39.28} & 16.790 & 3.931 & \textbf{22.143} & \textbf{16.402} & 12.788 & 33.122 & 33.404 & 18.933 & 20.914 \\
\bottomrule
\end{tabular}}
\caption{Analysis of training encoder along with the task model when fine-tuning. Experiments are conducted under the same setting to Sec.~\ref{main_result}}
\label{enc_train_table}
\end{table*}

Evaluations are done in the same settings as in Table~\ref{main_result_table}. By applying this two-step training method, we succeed in obtaining better performance than the baseline in most of the settings.  It can be inferred that compared with the setting of a frozen encoder (i.e., $\phi$ is not optimized), further introducing trainable encoder parameters did not further enhance the model's performance as anticipated. Although we can achieve better results in most settings than baseline, performance in held-in and held-out settings seems to be less stable compared to the "frozen encoder" setting. Particularly, we find that optimizing the encoder results in degraded performance in the ICL setting, especially after being fine-tuned on TriviaQA datasets. We attribute this to the fact that million-scale parameter models, after fine-tuning on certain data, cannot guarantee to generalize the encoding capability to a broader range of scenarios, e.g. the ICL setting, as defined in Table~\ref{data_format}. We present the results of the second training strategy discussed in Sec.~\ref{training_sec} in Table~\ref{enc_train_table}.

\subsection{ICL Setting w/o Contexts}
\label{icl_no_ctx_sec}

\begin{table}
\tabcolsep=0.12cm
\centering
\small
\scalebox{1.0}{
\begin{tabular}{l|cccc} 
\toprule
\multirow{2}{*}{\begin{tabular}[c]{@{}l@{}}\textbf{ICL samples}\\\textbf{w/o contexts}\end{tabular}} & \multicolumn{2}{c}{\textbf{TriviaQA}} & \multicolumn{2}{c}{\textbf{NQ}} \\
 & \textit{dev} & \textit{test} & \textit{dev} & \textit{test} \\ 
\midrule
\midrule
baseline & 20.052 & 20.083 & 19.242 & 19.529 \\
+ 10 vec & 19.939 & 20.233 & \textbf{19.539} & \textbf{19.668} \\
+ 20 vec & \textbf{20.358} & \textbf{20.578} & 19.333 & 19.501 \\
\bottomrule
\end{tabular}}
\caption{Result of fine-tuning on data with ICL samples (without context information) and evaluating on held-in setting.}
\label{icl_no_cxt_table}
\end{table}

We also experiment with optimizing objective \ref{our_objective_icl} defined in Sec.~\ref{icl_setting_sec} where only query-answer pairs are provided in the ICL format input. The detailed data format is shown in Table~\ref{data_format} "ICL format w/o contexts" and the query-answer pair is sampled as many as possible from the held-in dataset. The utility of the encoder remains the same as it encodes 10 (+ 10 vec) or 20 (+ 20 vec) pieces of context and is kept frozen during the training. 

The model is fine-tuned on TriviaQA and NQ and evaluated in held-in settings. We report the result in Table~\ref{icl_no_cxt_table}. First, we see that our method can still enhance the model in this setting but the improvements seem to be not consistent or prominent. Second, notice that the improvement on each dataset is not as remarkable as that in the ICL setting in Table~\ref{main_result_table}, where each ICL sample is provided along with one piece of context.

To summarize the findings here, our method for encoding context exhibits a more pronounced performance enhancement in ICL settings that incorporate context information. We posit that the underlying reason for this is that the cross-attention mechanism, which facilitates information interchange between inputs (embedded by the task model) and dense context information (encoded by the encoder), is particularly effective when context interacts with context, instead of context with ICL samples with only query-answer pairs.

\subsection{A More Challenging Setting}

\begin{table}
\tabcolsep=0.12cm
\centering
\small
\scalebox{1.0}{
\begin{tabular}{l|cccc} 
\toprule
\multirow{2}{*}{\begin{tabular}[c]{@{}l@{}}\textbf{$\bm{k_{max}}=\{\}$}\end{tabular}} & \multicolumn{2}{c}{\textbf{TriviaQA}} & \multicolumn{2}{c}{\textbf{NQ}} \\
 & \textit{dev} & \textit{test} & \textit{dev} & \textit{test} \\ 
\midrule
\midrule
baseline & 20.391 & 20.472 & 18.968 & 19.889 \\
+ 1 vec & 21.636 & 21.533 & 20.041 & 20.637 \\
+ 5 vec & 21.942 & 22.010 & 20.258 & 20.942 \\
+ 10 vec & \textbf{21.964} & \textbf{22.072} & \textbf{22.268} & \textbf{22.632} \\
\bottomrule
\end{tabular}}
\caption{Effectiveness of our method on encoding when we remove the influence on text form context information in $\bm{x}$. }
\label{hidden_align}
\end{table}

In our method presented in Sec.~\ref{enc_and_cros}, we adopt a projector module that is applied to align the high-dimensional hidden spaces and adopt cross-attention mechanism to incorporate the dense context information in each layer. In this section, we evaluate the effectiveness of our method in a more challenging setting. 

Specifically, compared to the data format stated in the "Held-in Held-out" setting in Table~\ref{data_format}, we remove the contexts in input $\bm{x}$ and keep only questions and answers in the training data, i.e., $\bm{x}$ in objective \ref{our_objective} becomes $(\bm{q}, \bm{a}, \{\}, \bm{P})$. Only several contexts are supplied as "Additional Contexts" encoded by the encoder. Note that though supplying text-form contexts can greatly enhance models in ODQA tasks, here we remove them to test the effectiveness of the encoder and cross-attention mechanism in a more challenging setting. 

Results are shown in Table.~\ref{hidden_align}. "+ 1/5/10 vec" means we utilize 1/5/10 pieces of contexts and encode them into 1/5/10 vectors by taking the \textit{[CLS]} tokens' hidden states. It can be inferred that, firstly, with only one encoded vector, our method can enhance the model. Secondly, we observe consistent improvement across two datasets and three variants of our method that incorporating more contexts leads to better performance.

\section{Related Work}

\subsection{Retrieval Augmentation}
Recently, retrieval augmentation has been utilized to improve a large amount of Natural Language Processing downstream tasks such as question-answering \cite{chen-etal-2017-reading, lewis2020retrieval, kwiatkowski-etal-2019-natural, fan-etal-2019-eli5}, dialogue \cite{moghe-etal-2018-towards}, language modeling \cite{khandelwal2020generalization}, NER \cite{wang-etal-2022-damo, wang2021improving} and machine translation \cite{gu2018search, xu-etal-2022-boosting}. In the aforementioned work, the utilization of retrieval information has been fundamentally capable of enhancing model performance across all dimensions.

\subsection{Related Model Architectures}
Referring to the base model, there has been increasing interest in using models of encoder-decoder or decoder-only architectures in solving downstream tasks with retrieval augmentation recently. 

\citet{allaouzi2019encoder} and \citet{zhou2023medical} employ models of encoder-decoder architectures to solve visual question answering task in the medical domain. In their work, the encoder model is responsible for extracting prominent features from a medical image and the decoder part generates the answer. \citet{math11071624} utilizes an encoder-decoder model with constrained decoding to solve extractive question answering task. 

Decoder-only models, e.g., ChatGPT and GPT-4 \cite{achiam2023gpt}, are more famous for their surprisingly great performance on tasks like question answering \cite{ali2022performance} and there is abundant work that tries to improve the performance based on GPTs \cite{pereira2023visconde}. \citet{kim2024rag} introduce a chatbot model that utilizes generative AI and the Retrieval Augmented Generation method to address the issue that achieving regulatory compliance necessitates the intricate navigation of exceptionally complex and voluminous guidelines in the pharmaceutical industry.

In our work, we also incorporate an encoder for context encoding. However, compared to the traditional encoder-decoder models, the encoder part in our method is several times smaller than the decoder part. Although our method does not alter the quadratic complexity of the attention mechanism, it instead processes the long contexts in a much lower dimension, thus being able to quintuple the capacity to cover context information without the need to utilize additional computing resources.

\subsection{Utilizing Long Contexts}

To handle contexts with excessive length, recently proposed techniques such as context compression are increasingly investigated in NLP research. 

\citet{chevalier2023adapting} proposes "AutoCompressors" that uses OPT \cite{zhang2022opt} and Llama-2 \cite{touvron2023llama} to compress texts into summary vectors and show that utilizing long contexts can improve perplexity. In their method, the compression is done by the billion-level language model, and in one of their experiments, they train on sequence with 30720 tokens with 20 compression steps. However, the complete computation graph cannot be fully kept in such settings, and the optimizing process has to rely on stopping gradients, which poses potential risks to the mathematical principle behind gradient descent. Similarly in \citet{zhang2024soaring}'s work, the long context is first partitioned into multiple intervals, and then a sliding window is employed to sequentially process one interval at a time and the compressed token embeddings are kept for the next token prediction. It is implemented by introducing additional trainable parameters to the origin language model to finish the task of "Activation Condensing", and original parameters are frozen throughout the training process.


\section{Conclusion}

In this paper, we propose a method that incorporates a small encoder model for excessively long context encoding by applying cross-attention mechanism with the original task model. The method is simple and general for transformer-based language models. In our experiments, after fine-tuning on ODQA dataset, we find improved performance across two held-in, four held-out and two ICL settings, compared to a baseline that incorporates the reranking technique on training data, showing the effectiveness of our method in utilizing long contexts. 
We note that the intuitive explanations for the performance improvement are as follows: 1) the encoder model provides the ability to encode longer contexts; 2) the cross-attention mechanism is useful in selectively attending the correct parts of the inputs.
Regarding the efficiency, the need for GPU quantity remains unchanged and the run time remains competitive to the baseline. 
\section{Limitations}
First, we have only tested our method in 1B7 models with a 110M encoder, and yet we have not tested the effectiveness of our method on larger language models, e.g., 7B and 70B, due to limited computing resources. 

Second, we observe that our method exhibits relatively modest performance under setting \ref{icl_no_ctx_sec}, with only a slight improvement compared to the baseline. We attribute the potential reasons for this to the cross-attention mechanism being unsuitable for modeling the relationship between context and ICL samples (without contexts).
\section*{Acknowledgement}
This work was supported by Alibaba Group through Alibaba Innovative Research Program.

\bibliography{custom}

\begin{thebibliography}{42}
\expandafter\ifx\csname natexlab\endcsname\relax\def\natexlab#1{#1}\fi

\bibitem[{Achiam et~al.(2023)Achiam, Adler, Agarwal, Ahmad, Akkaya, Aleman, Almeida, Altenschmidt, Altman, Anadkat et~al.}]{achiam2023gpt}
Josh Achiam, Steven Adler, Sandhini Agarwal, Lama Ahmad, Ilge Akkaya, Florencia~Leoni Aleman, Diogo Almeida, Janko Altenschmidt, Sam Altman, Shyamal Anadkat, et~al. 2023.
\newblock Gpt-4 technical report.
\newblock \emph{arXiv preprint arXiv:2303.08774}.

\bibitem[{Ali et~al.(2022)Ali, Tang, Connolly, Fridley, Shin, Sullivan, Cielo, Oyelese, Doberstein, Telfeian et~al.}]{ali2022performance}
Rohaid Ali, Oliver~Y Tang, Ian~D Connolly, Jared~S Fridley, John~H Shin, Patricia L~Zadnik Sullivan, Deus Cielo, Adetokunbo~A Oyelese, Curtis~E Doberstein, Albert~E Telfeian, et~al. 2022.
\newblock Performance of chatgpt, gpt-4, and google bard on a neurosurgery oral boards preparation question bank.
\newblock \emph{Neurosurgery}, pages 10--1227.

\bibitem[{Allaouzi et~al.(2019)Allaouzi, Ahmed, and Benamrou}]{allaouzi2019encoder}
Imane Allaouzi, Mohamed~Ben Ahmed, and Badr Benamrou. 2019.
\newblock An encoder-decoder model for visual question answering in the medical domain.
\newblock In \emph{CLEF (working notes)}.

\bibitem[{Berant et~al.(2013)Berant, Chou, Frostig, and Liang}]{berant-etal-2013-semantic}
Jonathan Berant, Andrew Chou, Roy Frostig, and Percy Liang. 2013.
\newblock \href {https://www.aclweb.org/anthology/D13-1160} {Semantic parsing on {F}reebase from question-answer pairs}.
\newblock In \emph{Proceedings of the 2013 Conference on Empirical Methods in Natural Language Processing}, pages 1533--1544, Seattle, Washington, USA. Association for Computational Linguistics.

\bibitem[{Brown et~al.(2020)Brown, Mann, Ryder, Subbiah, Kaplan, Dhariwal, Neelakantan, Shyam, Sastry, Askell et~al.}]{brown2020language}
Tom Brown, Benjamin Mann, Nick Ryder, Melanie Subbiah, Jared~D Kaplan, Prafulla Dhariwal, Arvind Neelakantan, Pranav Shyam, Girish Sastry, Amanda Askell, et~al. 2020.
\newblock Language models are few-shot learners.
\newblock \emph{Advances in neural information processing systems}, 33:1877--1901.

\bibitem[{Chen et~al.(2017)Chen, Fisch, Weston, and Bordes}]{chen-etal-2017-reading}
Danqi Chen, Adam Fisch, Jason Weston, and Antoine Bordes. 2017.
\newblock \href {https://doi.org/10.18653/v1/P17-1171} {Reading {W}ikipedia to answer open-domain questions}.
\newblock In \emph{Proceedings of the 55th Annual Meeting of the Association for Computational Linguistics (Volume 1: Long Papers)}, pages 1870--1879, Vancouver, Canada. Association for Computational Linguistics.

\bibitem[{Chevalier et~al.(2023)Chevalier, Wettig, Ajith, and Chen}]{chevalier2023adapting}
Alexis Chevalier, Alexander Wettig, Anirudh Ajith, and Danqi Chen. 2023.
\newblock Adapting language models to compress contexts.
\newblock \emph{arXiv preprint 2305.14788}.

\bibitem[{Chowdhery et~al.(2023)Chowdhery, Narang, Devlin, Bosma, Mishra, Roberts, Barham, Chung, Sutton, Gehrmann et~al.}]{chowdhery2023palm}
Aakanksha Chowdhery, Sharan Narang, Jacob Devlin, Maarten Bosma, Gaurav Mishra, Adam Roberts, Paul Barham, Hyung~Won Chung, Charles Sutton, Sebastian Gehrmann, et~al. 2023.
\newblock Palm: Scaling language modeling with pathways.
\newblock \emph{Journal of Machine Learning Research}, 24(240):1--113.

\bibitem[{Devlin et~al.(2018)Devlin, Chang, Lee, and Toutanova}]{devlin2018bert}
Jacob Devlin, Ming-Wei Chang, Kenton Lee, and Kristina Toutanova. 2018.
\newblock Bert: Pre-training of deep bidirectional transformers for language understanding.
\newblock \emph{arXiv preprint arXiv:1810.04805}.

\bibitem[{Dong et~al.(2022)Dong, Li, Dai, Zheng, Wu, Chang, Sun, Xu, and Sui}]{dong2022survey}
Qingxiu Dong, Lei Li, Damai Dai, Ce~Zheng, Zhiyong Wu, Baobao Chang, Xu~Sun, Jingjing Xu, and Zhifang Sui. 2022.
\newblock A survey for in-context learning.
\newblock \emph{arXiv preprint arXiv:2301.00234}.

\bibitem[{Fan et~al.(2019)Fan, Jernite, Perez, Grangier, Weston, and Auli}]{fan-etal-2019-eli5}
Angela Fan, Yacine Jernite, Ethan Perez, David Grangier, Jason Weston, and Michael Auli. 2019.
\newblock \href {https://doi.org/10.18653/v1/P19-1346} {{ELI}5: Long form question answering}.
\newblock In \emph{Proceedings of the 57th Annual Meeting of the Association for Computational Linguistics}, pages 3558--3567, Florence, Italy. Association for Computational Linguistics.

\bibitem[{Gu et~al.(2018)Gu, Wang, Cho, and Li}]{gu2018search}
Jiatao Gu, Yong Wang, Kyunghyun Cho, and Victor~OK Li. 2018.
\newblock Search engine guided neural machine translation.
\newblock In \emph{Proceedings of the AAAI Conference on Artificial Intelligence}, volume~32.

\bibitem[{Izacard et~al.(2021)Izacard, Caron, Hosseini, Riedel, Bojanowski, Joulin, and Grave}]{izacard2021unsupervised}
Gautier Izacard, Mathilde Caron, Lucas Hosseini, Sebastian Riedel, Piotr Bojanowski, Armand Joulin, and Edouard Grave. 2021.
\newblock Unsupervised dense information retrieval with contrastive learning.
\newblock \emph{arXiv preprint arXiv:2112.09118}.

\bibitem[{Izacard and Grave(2020)}]{izacard2020leveraging}
Gautier Izacard and Edouard Grave. 2020.
\newblock Leveraging passage retrieval with generative models for open domain question answering.
\newblock \emph{arXiv preprint arXiv:2007.01282}.

\bibitem[{Izacard et~al.(2022)Izacard, Lewis, Lomeli, Hosseini, Petroni, Schick, Dwivedi-Yu, Joulin, Riedel, and Grave}]{izacard2022few}
Gautier Izacard, Patrick Lewis, Maria Lomeli, Lucas Hosseini, Fabio Petroni, Timo Schick, Jane Dwivedi-Yu, Armand Joulin, Sebastian Riedel, and Edouard Grave. 2022.
\newblock Few-shot learning with retrieval augmented language models.
\newblock \emph{arXiv preprint arXiv:2208.03299}.

\bibitem[{Joshi et~al.(2017)Joshi, Choi, Weld, and Zettlemoyer}]{joshi-etal-2017-triviaqa}
Mandar Joshi, Eunsol Choi, Daniel Weld, and Luke Zettlemoyer. 2017.
\newblock \href {https://doi.org/10.18653/v1/P17-1147} {{T}rivia{QA}: A large scale distantly supervised challenge dataset for reading comprehension}.
\newblock In \emph{Proceedings of the 55th Annual Meeting of the Association for Computational Linguistics (Volume 1: Long Papers)}, pages 1601--1611, Vancouver, Canada. Association for Computational Linguistics.

\bibitem[{Karpukhin et~al.(2020)Karpukhin, Oguz, Min, Lewis, Wu, Edunov, Chen, and Yih}]{karpukhin2020dense}
Vladimir Karpukhin, Barlas Oguz, Sewon Min, Patrick Lewis, Ledell Wu, Sergey Edunov, Danqi Chen, and Wen-tau Yih. 2020.
\newblock Dense passage retrieval for open-domain question answering.
\newblock In \emph{Proceedings of the 2020 Conference on Empirical Methods in Natural Language Processing (EMNLP)}. Association for Computational Linguistics.

\bibitem[{Khalifa et~al.(2023)Khalifa, Logeswaran, Lee, Lee, and Wang}]{khalifa-etal-2023-shot}
Muhammad Khalifa, Lajanugen Logeswaran, Moontae Lee, Honglak Lee, and Lu~Wang. 2023.
\newblock \href {https://doi.org/10.18653/v1/2023.acl-long.885} {Few-shot reranking for multi-hop {QA} via language model prompting}.
\newblock In \emph{Proceedings of the 61st Annual Meeting of the Association for Computational Linguistics (Volume 1: Long Papers)}, pages 15882--15897, Toronto, Canada. Association for Computational Linguistics.

\bibitem[{Khandelwal et~al.(2020)Khandelwal, Levy, Jurafsky, Zettlemoyer, and Lewis}]{khandelwal2020generalization}
Urvashi Khandelwal, Omer Levy, Dan Jurafsky, Luke Zettlemoyer, and Mike Lewis. 2020.
\newblock \href {http://arxiv.org/abs/1911.00172} {Generalization through memorization: Nearest neighbor language models}.

\bibitem[{Kim et~al.(2022)Kim, Cho, Kim, Kim, Yoo, and Lee}]{kim2022self}
Hyuhng~Joon Kim, Hyunsoo Cho, Junyeob Kim, Taeuk Kim, Kang~Min Yoo, and Sang-goo Lee. 2022.
\newblock Self-generated in-context learning: Leveraging auto-regressive language models as a demonstration generator.
\newblock \emph{arXiv preprint arXiv:2206.08082}.

\bibitem[{Kim and Min(2024)}]{kim2024rag}
Jaewoong Kim and Moohong Min. 2024.
\newblock From rag to qa-rag: Integrating generative ai for pharmaceutical regulatory compliance process.
\newblock \emph{arXiv preprint arXiv:2402.01717}.

\bibitem[{Kwiatkowski et~al.(2019)Kwiatkowski, Palomaki, Redfield, Collins, Parikh, Alberti, Epstein, Polosukhin, Devlin, Lee, Toutanova, Jones, Kelcey, Chang, Dai, Uszkoreit, Le, and Petrov}]{kwiatkowski-etal-2019-natural}
Tom Kwiatkowski, Jennimaria Palomaki, Olivia Redfield, Michael Collins, Ankur Parikh, Chris Alberti, Danielle Epstein, Illia Polosukhin, Jacob Devlin, Kenton Lee, Kristina Toutanova, Llion Jones, Matthew Kelcey, Ming-Wei Chang, Andrew~M. Dai, Jakob Uszkoreit, Quoc Le, and Slav Petrov. 2019.
\newblock \href {https://doi.org/10.1162/tacl_a_00276} {Natural questions: A benchmark for question answering research}.
\newblock \emph{Transactions of the Association for Computational Linguistics}, 7:452--466.

\bibitem[{Lewis et~al.(2020)Lewis, Perez, Piktus, Petroni, Karpukhin, Goyal, K{\"u}ttler, Lewis, Yih, Rockt{\"a}schel et~al.}]{lewis2020retrieval}
Patrick Lewis, Ethan Perez, Aleksandra Piktus, Fabio Petroni, Vladimir Karpukhin, Naman Goyal, Heinrich K{\"u}ttler, Mike Lewis, Wen-tau Yih, Tim Rockt{\"a}schel, et~al. 2020.
\newblock Retrieval-augmented generation for knowledge-intensive nlp tasks.
\newblock \emph{Advances in Neural Information Processing Systems}, 33:9459--9474.

\bibitem[{Li et~al.(2023)Li, Sun, Liu, Liu, and Ji}]{math11071624}
Shaobo Li, Chengjie Sun, Bingquan Liu, Yuanchao Liu, and Zhenzhou Ji. 2023.
\newblock \href {https://doi.org/10.3390/math11071624} {Modeling extractive question answering using encoder-decoder models with constrained decoding and evaluation-based reinforcement learning}.
\newblock \emph{Mathematics}, 11(7).

\bibitem[{Moghe et~al.(2018)Moghe, Arora, Banerjee, and Khapra}]{moghe-etal-2018-towards}
Nikita Moghe, Siddhartha Arora, Suman Banerjee, and Mitesh~M. Khapra. 2018.
\newblock \href {https://doi.org/10.18653/v1/D18-1255} {Towards exploiting background knowledge for building conversation systems}.
\newblock In \emph{Proceedings of the 2018 Conference on Empirical Methods in Natural Language Processing}, pages 2322--2332, Brussels, Belgium. Association for Computational Linguistics.

\bibitem[{Muennighoff et~al.(2022)Muennighoff, Wang, Sutawika, Roberts, Biderman, Scao, Bari, Shen, Yong, Schoelkopf et~al.}]{muennighoff2022crosslingual}
Niklas Muennighoff, Thomas Wang, Lintang Sutawika, Adam Roberts, Stella Biderman, Teven~Le Scao, M~Saiful Bari, Sheng Shen, Zheng-Xin Yong, Hailey Schoelkopf, et~al. 2022.
\newblock Crosslingual generalization through multitask finetuning.
\newblock \emph{arXiv preprint arXiv:2211.01786}.

\bibitem[{Pereira et~al.(2023)Pereira, Fidalgo, Lotufo, and Nogueira}]{pereira2023visconde}
Jayr Pereira, Robson Fidalgo, Roberto Lotufo, and Rodrigo Nogueira. 2023.
\newblock Visconde: Multi-document qa with gpt-3 and neural reranking.
\newblock In \emph{European Conference on Information Retrieval}, pages 534--543. Springer.

\bibitem[{Radford et~al.(2019)Radford, Wu, Child, Luan, Amodei, Sutskever et~al.}]{radford2019language}
Alec Radford, Jeffrey Wu, Rewon Child, David Luan, Dario Amodei, Ilya Sutskever, et~al. 2019.
\newblock Language models are unsupervised multitask learners.
\newblock \emph{OpenAI blog}, 1(8):9.

\bibitem[{Rajpurkar et~al.(2016)Rajpurkar, Zhang, Lopyrev, and Liang}]{rajpurkar2016squad}
Pranav Rajpurkar, Jian Zhang, Konstantin Lopyrev, and Percy Liang. 2016.
\newblock Squad: 100,000+ questions for machine comprehension of text.
\newblock In \emph{Proceedings of the 2016 Conference on Empirical Methods in Natural Language Processing}, pages 2383--2392.

\bibitem[{Robertson et~al.(2009)Robertson, Zaragoza et~al.}]{robertson2009probabilistic}
Stephen Robertson, Hugo Zaragoza, et~al. 2009.
\newblock The probabilistic relevance framework: Bm25 and beyond.
\newblock \emph{Foundations and Trends{\textregistered} in Information Retrieval}, 3(4):333--389.

\bibitem[{Shi et~al.(2023)Shi, Min, Yasunaga, Seo, James, Lewis, Zettlemoyer, and Yih}]{shi2023replug}
Weijia Shi, Sewon Min, Michihiro Yasunaga, Minjoon Seo, Rich James, Mike Lewis, Luke Zettlemoyer, and Wen-tau Yih. 2023.
\newblock Replug: Retrieval-augmented black-box language models.
\newblock \emph{arXiv preprint arXiv:2301.12652}.

\bibitem[{Talmor and Berant(2018)}]{talmor2018web}
Alon Talmor and Jonathan Berant. 2018.
\newblock \href {http://arxiv.org/abs/1803.06643} {The web as a knowledge-base for answering complex questions}.

\bibitem[{Talmor et~al.(2019)Talmor, Herzig, Lourie, and Berant}]{talmor2019commonsenseqa}
Alon Talmor, Jonathan Herzig, Nicholas Lourie, and Jonathan Berant. 2019.
\newblock Commonsenseqa: A question answering challenge targeting commonsense knowledge.
\newblock In \emph{Proceedings of the 2019 Conference of the North American Chapter of the Association for Computational Linguistics: Human Language Technologies, Volume 1 (Long and Short Papers)}, pages 4149--4158.

\bibitem[{Touvron et~al.(2023)Touvron, Lavril, Izacard, Martinet, Lachaux, Lacroix, Rozi{\`e}re, Goyal, Hambro, Azhar et~al.}]{touvron2023llama}
Hugo Touvron, Thibaut Lavril, Gautier Izacard, Xavier Martinet, Marie-Anne Lachaux, Timoth{\'e}e Lacroix, Baptiste Rozi{\`e}re, Naman Goyal, Eric Hambro, Faisal Azhar, et~al. 2023.
\newblock Llama: Open and efficient foundation language models.
\newblock \emph{arXiv preprint arXiv:2302.13971}.

\bibitem[{Vaswani et~al.(2017)Vaswani, Shazeer, Parmar, Uszkoreit, Jones, Gomez, Kaiser, and Polosukhin}]{vaswani2017attention}
Ashish Vaswani, Noam Shazeer, Niki Parmar, Jakob Uszkoreit, Llion Jones, Aidan~N Gomez, {\L}ukasz Kaiser, and Illia Polosukhin. 2017.
\newblock Attention is all you need.
\newblock \emph{Advances in neural information processing systems}, 30.

\bibitem[{Wang et~al.(2021)Wang, Jiang, Bach, Wang, Huang, Huang, and Tu}]{wang2021improving}
Xinyu Wang, Yong Jiang, Nguyen Bach, Tao Wang, Zhongqiang Huang, Fei Huang, and Kewei Tu. 2021.
\newblock {{Improving Named Entity Recognition by External Context Retrieving and Cooperative Learning}}.
\newblock In \emph{{the Joint Conference of the 59th Annual Meeting of the Association for Computational Linguistics and the 11th International Joint Conference on Natural Language Processing (\textbf{ACL-IJCNLP 2021})}}. Association for Computational Linguistics.

\bibitem[{Wang et~al.(2022)Wang, Shen, Cai, Wang, Wang, Xie, Huang, Lu, Zhuang, Tu, Lu, and Jiang}]{wang-etal-2022-damo}
Xinyu Wang, Yongliang Shen, Jiong Cai, Tao Wang, Xiaobin Wang, Pengjun Xie, Fei Huang, Weiming Lu, Yueting Zhuang, Kewei Tu, Wei Lu, and Yong Jiang. 2022.
\newblock \href {https://doi.org/10.18653/v1/2022.semeval-1.200} {{DAMO}-{NLP} at {S}em{E}val-2022 task 11: A knowledge-based system for multilingual named entity recognition}.
\newblock In \emph{Proceedings of the 16th International Workshop on Semantic Evaluation (SemEval-2022)}, pages 1457--1468, Seattle, United States. Association for Computational Linguistics.

\bibitem[{Workshop et~al.(2022)Workshop, Scao, Fan, Akiki, Pavlick, Ili{\'c}, Hesslow, Castagn{\'e}, Luccioni, Yvon et~al.}]{workshop2022bloom}
BigScience Workshop, Teven~Le Scao, Angela Fan, Christopher Akiki, Ellie Pavlick, Suzana Ili{\'c}, Daniel Hesslow, Roman Castagn{\'e}, Alexandra~Sasha Luccioni, Fran{\c{c}}ois Yvon, et~al. 2022.
\newblock Bloom: A 176b-parameter open-access multilingual language model.
\newblock \emph{arXiv preprint arXiv:2211.05100}.

\bibitem[{Xu et~al.(2022)Xu, Crego, and Senellart}]{xu-etal-2022-boosting}
Jitao Xu, Josep Crego, and Jean Senellart. 2022.
\newblock \href {https://aclanthology.org/2022.amta-upg.20} {Boosting neural machine translation with similar translations}.
\newblock In \emph{Proceedings of the 15th Biennial Conference of the Association for Machine Translation in the Americas (Volume 2: Users and Providers Track and Government Track)}, pages 282--292, Orlando, USA. Association for Machine Translation in the Americas.

\bibitem[{Zhang et~al.(2024)Zhang, Liu, Xiao, Shao, Ye, and Dou}]{zhang2024soaring}
Peitian Zhang, Zheng Liu, Shitao Xiao, Ninglu Shao, Qiwei Ye, and Zhicheng Dou. 2024.
\newblock Soaring from 4k to 400k: Extending llm's context with activation beacon.
\newblock \emph{arXiv preprint arXiv:2401.03462}.

\bibitem[{Zhang et~al.(2022)Zhang, Roller, Goyal, Artetxe, Chen, Chen, Dewan, Diab, Li, Lin, Mihaylov, Ott, Shleifer, Shuster, Simig, Koura, Sridhar, Wang, and Zettlemoyer}]{zhang2022opt}
Susan Zhang, Stephen Roller, Naman Goyal, Mikel Artetxe, Moya Chen, Shuohui Chen, Christopher Dewan, Mona Diab, Xian Li, Xi~Victoria Lin, Todor Mihaylov, Myle Ott, Sam Shleifer, Kurt Shuster, Daniel Simig, Punit~Singh Koura, Anjali Sridhar, Tianlu Wang, and Luke Zettlemoyer. 2022.
\newblock \href {http://arxiv.org/abs/2205.01068} {Opt: Open pre-trained transformer language models}.

\bibitem[{Zhou et~al.(2023)Zhou, Mei, Yu, and Syeda-Mahmood}]{zhou2023medical}
Yuan Zhou, Jing Mei, Yiqin Yu, and Tanveer Syeda-Mahmood. 2023.
\newblock Medical visual question answering using joint self-supervised learning.
\newblock \emph{arXiv preprint arXiv:2302.13069}.

\end{thebibliography}
\bibliographystyle{acl_natbib}

\appendix
\section{Appendix}
\label{appendix}




\subsection{CommonsenseQA Format}
\label{comqa_app}

We show how we reformat data from CommonsenseQA in Table~\ref{comqa_table}. Reformated \textit{choice} turn A/B/C/D/E into 1/2/3/4/5 to avoid causing ambiguity with “A:” in prompts $\bm{P}$. The choices are removed in \textit{seq2seq} format and the problem becomes more challenging.

\begin{table*}[ht]
\centering
\scalebox{1.0}{
\begin{tabular}{ll}
\toprule
\multicolumn{1}{c}{\textbf{Setting}} & \multicolumn{1}{c}{\textbf{Format}} \\
\midrule
\midrule
Origin Format & \begin{tabular}[c]{@{}l@{}}A revolving door is convenient for two direction travel, but it also \\serves as a security measure at a what?\\A: bank\\B: library\\C: department store\\D: mall\\E: new york\\\\Answer: A\end{tabular} \\
\midrule
Reformatted \textit{choice} & \begin{tabular}[c]{@{}l@{}}Q: A revolving door is convenient for two direction travel, but it also \\serves as a security measure at a what? Choose from 1-5 given below.\\1: bank\\2: library\\3: department store\\4: mall\\5: new york\\A:~\\\\Answer: 1 or bank\end{tabular} \\
\midrule
Reformatted~\textit{seq2seq} & \begin{tabular}[c]{@{}l@{}}Q: A revolving door is convenient for two direction travel, but it also \\serves as a security measure at a what?\\A:~\\\\Answer: bank\end{tabular} \\
\bottomrule
\end{tabular}}
\caption{}
\label{comqa_table}
\end{table*}

\end{document}